\documentclass[final]{cvpr}

\usepackage{times}
\usepackage{epsfig}
\usepackage{graphicx}
\usepackage{amsmath}
\usepackage{amssymb}
\usepackage[linesnumbered,ruled,vlined]{algorithm2e}
\usepackage{algpseudocode}
\usepackage{amsmath}
\usepackage{bm}
\usepackage{color}
\usepackage{makecell}
\usepackage{amsmath}
\usepackage{amsmath,subfigure,epsfig,bbding}
\usepackage[tableposition=top]{caption}
\usepackage[figuresright]{rotating}
\usepackage{threeparttable}
\usepackage{multirow}
\usepackage{booktabs}
\newcommand{\PreserveBackslash}[1]{\let\temp=\\#1\let\\=\temp}
\newcolumntype{C}[1]{>{\PreserveBackslash\centering}p{#1}}
\newcolumntype{R}[1]{>{\PreserveBackslash\raggedleft}p{#1}}
\newcolumntype{L}[1]{>{\PreserveBackslash\raggedright}p{#1}}

\usepackage{array}
    \makeatletter
    \newcommand{\thickhline}{%
        \noalign {\ifnum 0=`}\fi \hrule height 1pt
        \futurelet \reserved@a \@xhline
}
\newcolumntype{"}{@{\vrule width 1pt}}
\makeatother

\usepackage[pagebackref=true,breaklinks=true,colorlinks,bookmarks=false]{hyperref}

\def\cvprPaperID{3890} 
\def\confYear{CVPR 2022}

\makeatletter
\def\thanks#1{\protected@xdef\@thanks{\@thanks
		\protect\footnotetext{#1}}}
\makeatother

\begin{document}

\title{Head and Body: Unified Detector and Graph Network \\for Person Search in Media}
\author{
	Xiujun Shu\textsuperscript{*},
	Yusheng Tao\textsuperscript{*}, 
	Ruizhi Qiao, 
	Bo Ke, 
	Wei Wen, 
	Bo Ren \\ 
	\small
	Tencent Youtu Lab.\\
	\thanks{indicates equal contribution of this work.}  
}
 

\maketitle

\begin{abstract}
Person search in media has seen increasing potential in Internet applications, such as video clipping and character collection. This task is common but overlooked by previous person search works which focus on surveillance scenes. The media scenarios have some different challenges from surveillance scenes. For example, a person may change his clothes frequently. To alleviate this issue, this paper proposes a \textbf{U}nified \textbf{D}etector and \textbf{G}raph \textbf{Net}work (\textbf{UDGNet}) for person search in media. UDGNet is the first person search framework to detect and re-identify the human body and head simultaneously. Specifically, it first builds two branches based on a unified network to detect the human body and head, then the detected body and head are used for re-identification. This dual-task approach can significantly enhance discriminative learning. To tackle the cloth-changing issue, UDGNet builds two graphs to explore reliable links among cloth-changing samples and utilizes a graph network to learn better embeddings. This design effectively enhances the robustness of person search to cloth-changing challenges. Besides, we demonstrate that UDGNet can be implemented with both anchor-based and anchor-free person search frameworks and further achieve performance improvement. This paper also contributes a large-scale dataset for \textbf{P}erson \textbf{S}earch in \textbf{M}edia (\textbf{PSM}), which provides both body and head annotations. It is by far the largest dataset for person search in media. Experiments show that UDGNet improves the anchor-free model AlignPS by 12.1\% in mAP. Meanwhile, it shows good generalization across surveillance and long-term scenarios. The dataset and code will be available at : \textcolor{magenta}{\url{https://github.com/shuxjweb/PSM.git}}.

\end{abstract}

\section{Introduction}

\begin{figure}[t]
	\centering
	\includegraphics[width=\linewidth,height=8.5cm]{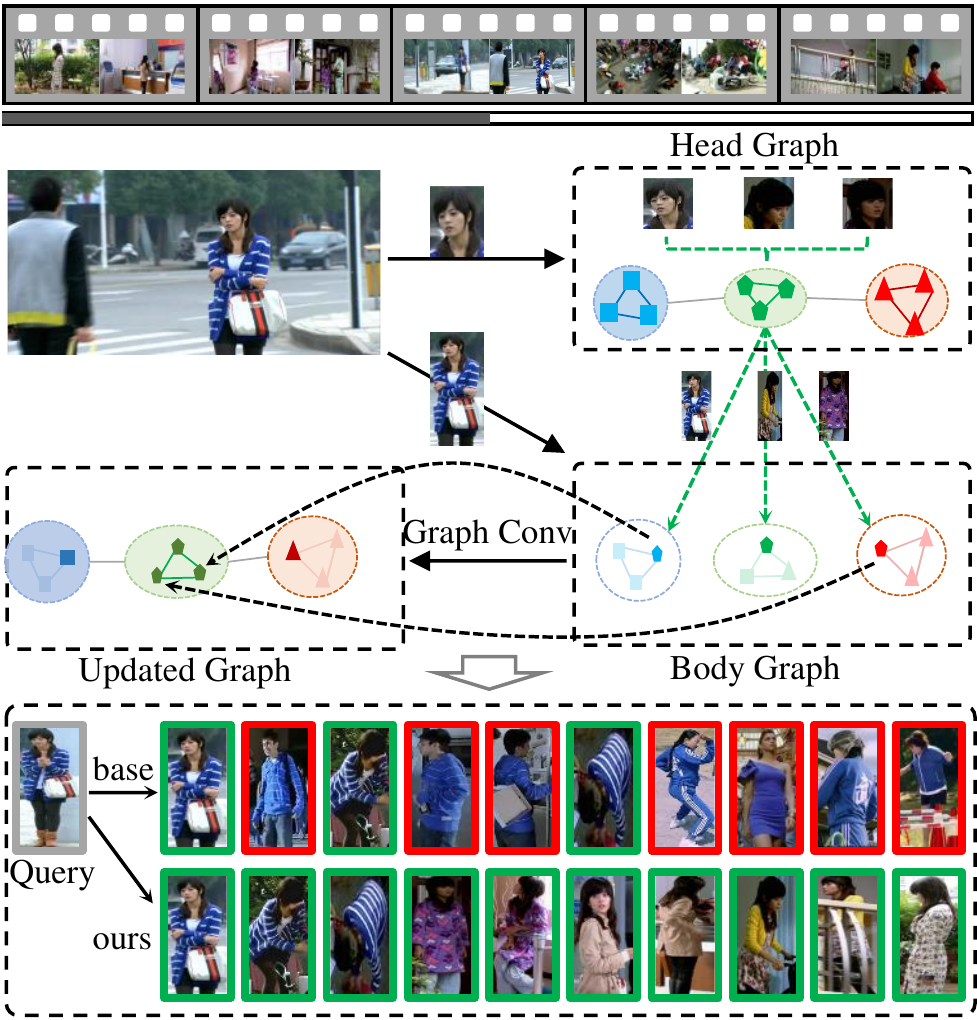}
	\caption{\textbf{Illustration of the key idea of UDGNet.}
		Compared with the body graph, the head graph could establish more reliable links among cloth-changing samples. It will help recall more positive samples and significantly improve the mean average precision.
	}
	\label{fig:motivation}
\end{figure}

Person search aims to find a probe person from a gallery of whole scene images \cite{xiao2017joint, zheng2017person}. It contains two sub-tasks: pedestrian detection and person re-identification (re-ID). Conventional methods deal with detection and re-ID separately, but person search jointly optimizes them in a unified framework. It presents more advantages in efficiency and has attracted tremendous attention in past years \cite{lan2018person, han2019re, zhong2020robust, wang2020tcts, kim2021prototype}. However, current works mainly focus on surveillance scenes and ignore the media scenarios. With the development of mobile Internet, video recreation has become more and more popular, \emph{e.g.,} video clipping and character collection. For example, to collect images or videos about a celebrity who appears at different activities. Person search is the key technique in these applications. 

In media, we need not only to address common challenges, \emph{e.g.,} occlusion, viewpoint variations, and background clutter, but also have to deal with the cloth-changing issue. Pedestrians in media change clothes far more often than in surveillance scenes. As shown in Fig.~\ref{fig:motivation}, the girl wears different clothes in the community, indoors, on the street, and in the hospital. It is very challenging to retrieve these cloth-changing samples. Existing person search works mainly focus on several aspects, \emph{e.g.,} identity-driven detection \cite{han2019re, wang2020tcts}, multi-scale learning \cite{lan2018person, jing2020pose}, and partial matching \cite{chen2018person, zhong2020robust}. Although several works \cite{yu2020cocas, yang2019person, shu2021semantic} in the re-ID task have made early exploration on the cloth-changing issue, it has not been studied fully to date. Moreover, few works have studied this problem in person search, especially the media scenes. 

This paper targets to design a unified person search framework robust to the cloth-changing issue in media. As shown in Fig.~\ref{fig:motivation}, the core idea is to establish more reliable links among cloth-changing samples. This idea leads to the Unified Detector and Graph Network (\textbf{UDGNet}) for person search in media. UDGNet consists of two modules: unifying head and body (UHB) network and head-driven graph (HDG) network. The UHB module aims to simultaneously detect and re-identify the human body and head based on the same network. 
Two open questions are naturally thrown at us: 1) Is it possible to detect body and head accurately based on the same backbone network? 2) Can the head re-ID help improve the final performance? The answers are yes. 
Extensive experiments verify its effectiveness in both detection and re-ID tasks. The HDG module aims to tackle the cloth-changing issue. As body features are sensitive to changed clothes, the head graph is employed to establish reliable links among cloth-changing samples.
Benefiting from this design, the distances among positive cloth-changing samples are shortened, yielding a higher recall rate.

To test our approach, we contribute a large-scale dataset named Person Search in Media (\textbf{PSM}). Compared with previous person search datasets, \emph{i.e.,} PRW \cite{zheng2017person} and CUHK-SYSU \cite{xiao2017joint}, PSM presents several characteristics:
1) Larger scale; 2) Accurate body and head annotations; 3) Collected in media scenes with new challenges. 
Experimental results show that UDGNet can achieve significant performance improvement in media scenes. To demonstrate its generalization, extensive experiments have been conducted on surveillance datasets, media datasets, and cloth-changing datasets. Consistent improvements have been seen in these experiments. To the best of our knowledge, UDGNet is the first work in person search to simultaneously detect and re-identify body and head based on the same network. It is also a pioneering attempt to study the cloth-changing issue in person search in media. 
We believe that the proposed methods and the contributed PSM could foster more future studies in person search in media. In summary, our major contributions are as follows:
\begin{itemize}
\item We propose a unifying head and body network, which could detect and re-identify body and head simultaneously. The dual-task framework enhances discriminative learning significantly. 
\item We propose a head-driven graph network to address the cloth-changing issue in person search in media. This is a pioneering work to establish reliable links via a head graph among cloth-changing samples. 
\item We contribute a large-scale dataset PSM for person search in media. PSM provides some new challenges and inspires more research efforts in media scenarios.
\end{itemize}

\begin{figure*}[t]
	\centering  
	\includegraphics[width=\linewidth]{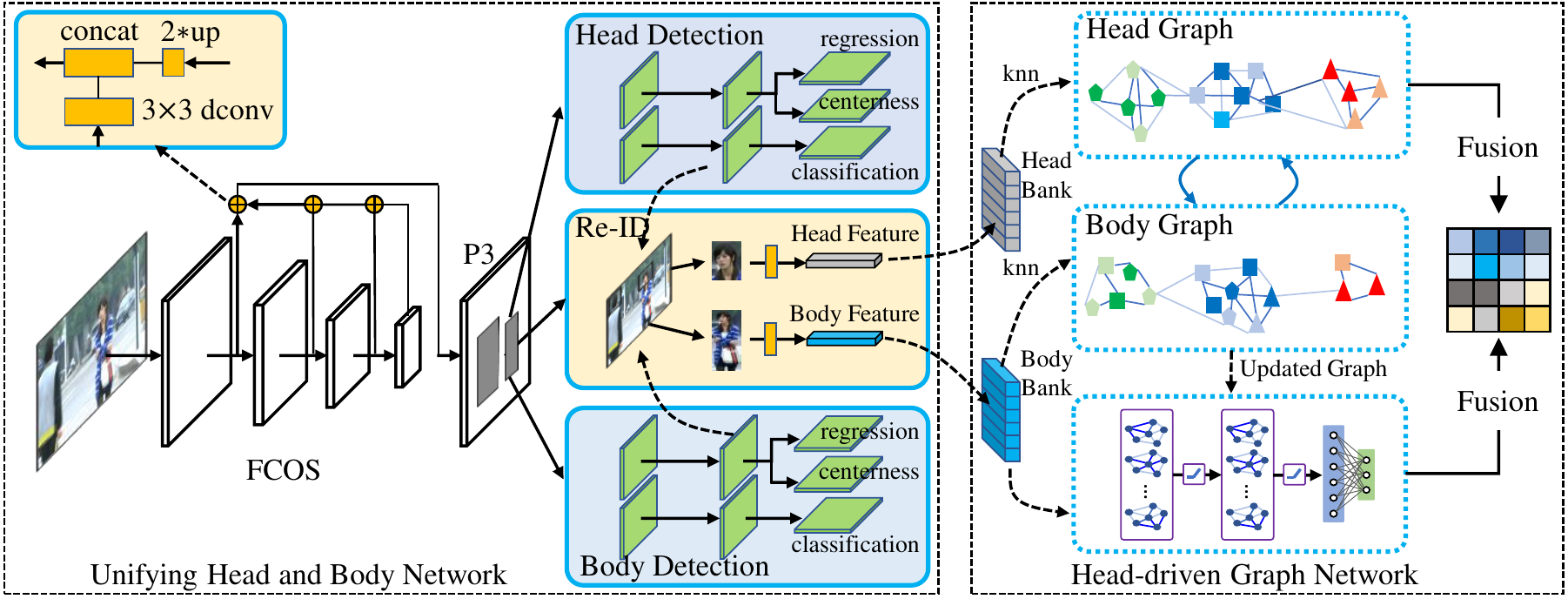}  
	\caption{\textbf{Architecture of the proposed UDGNet Framework.} 
		The framework consists of a unifying head and body (UHB) network and a head-driven graph (HDG) network. The UHB module contains head/body detection and re-identification. Its backbone follows the basic structure of FCOS \cite{tian2019fcos}. The HDG module builds two graphs and employs a graph network to learn better embeddings. Finally, the head and updated body features are fused with a strategy based on similarity matrices.}	
	\label{fig:framework}
\end{figure*}

\section{Related Work}
\subsection{Pedestrian Detection}
Pedestrian detection is an important sub-task in person search. Current detectors mainly follow the general object detection \cite{zhao2019object} which can be divided into two categories: two-stage detection \cite{ren2015faster, cai2018cascade, lin2017feature, song2020revisiting} and one-stage detection \cite{redmon2016you, liu2016ssd, tian2019fcos}. Most current person search works utilize Faster-RCNN due to its higher detection precision. Object detectors can also be classified into anchor-based and anchor-free ones. Anchor-free detectors \cite{law2018cornernet, yang2019reppoints} have attracted lots of attention recently due to their simple structures and efficient implementation. AlignPS \cite{yan2021anchor} first applies anchor-free detector into person search. In this work, we develop a unified framework to conduct body/head detection and re-identification. This framework can be implemented with both anchor-based and anchor-free approaches and further achieve performance improvement.

\subsection{Cloth-changing Person Re-identification}
Cloth-changing person re-ID is a new rising challenging research topic that aims at retrieving pedestrians whose clothes are changed. Several works \cite{huang2019beyond, yang2019person, yu2020cocas, qian2020long, hong2021fine, shu2021large} have contributed to this field. Most of them focus on body shape or contour sketch, but these cues are not robust enough to view and posture variations. The cloth-changing issue is much common in media than in surveillance systems. However, few works have discussed it in person search, especially the media scenes. This work fills the gap and provides an efficient solution.


\subsection{Graph Networks}

 Graph convolutional networks (GCNs) are effective for graph structure data \cite{defferrard2016convolutional, niepert2016learning} and have been applied to computer vision tasks \cite{yan2018spatial}. Several works \cite{shen2018person, zhou2019robust, wu2019unsupervised} have attempted to use GCN in the person re-ID task. Current works mainly focus on part-wise matching \cite{wu2020adaptive, yang2020spatial, wang2020high, zhang2021person} and context relation \cite{bao2019masked, ji2020context, yan2019learning}. The former exploits relations among body parts and the latter models the relation between the query and gallery sets to obtain optimal representation. However, these methods build graphs based on single-modal features only.
 That means they could only establish links among the same clothes. 
 Our method belongs to the latter but we propose an innovative graph construction to establish reliable links among cloth-changing samples.


\subsection{Person Search in Surveillance}
Person search was first proposed in \cite{xu2014person}. Two popular datasets are PRW \cite{zheng2017person} and CUHK-SYSU \cite{xiao2017joint}. Most person search works have been done with them. 
Current works can be divided into two-step and one-step methods. 
The two-step category \cite{zheng2017person, chen2018person, han2019re} conducts detection and re-identification separately. They first detect pedestrians and then crop the detected person for re-ID. The one-step category \cite{xiao2017joint, dong2020bi, chen2020norm} integrates detection and re-ID into the same network. Current works focus on identity-driven detection \cite{wang2020tcts}, multi-scale learning \cite{lan2018person, jing2020pose}, and partial matching \cite{chen2018person, zhong2020robust}. Most of them utilize Faster-RCNN \cite{ren2015faster} and OIM loss \cite{xiao2017joint} to enable end-to-end training.



\subsection{Person Search in Media}\label{2.5}
The media scenario has not been fully studied to date. Two related datasets are CSM \cite{ huang2018unifying} and MovieNet \cite{huang2020movienet}. 
However, the task they defined is a little different from person search. For example, the query images they provided are portraits, not complete human bodies, leading to the unavailability of body cues. The training samples are tracklets and they just need to locate the frames at which the target person appears. In this work, we provide an image dataset for pure person search in media which has a larger 
scale than previous ones.

\section{Methodology}

\subsection{Framework Overview}
In this section, we present our Unified Detector and Graph Network (UDGNet) for person search in media. 
As shown in Fig.~\ref{fig:framework}, UDGNet consists of two modules: unifying head and body (UHB) network and head-driven graph (HDG) network. The UHB module first detects head and body based on the same network and then conducts head and body re-ID, respectively. This is one of the key components of our framework, which can significantly enhance discriminative learning. The HDG module first employs the extracted features to build the head and body graphs. Then it utilizes a graph network to learn better representations. In general, body features have strong relationships with clothes. They are more sensitive to changing clothes than head features. That is to say, the head graph could establish more reliable links among cloth-changing samples. This is another crucial component of our framework. 
Finally, the head and updated body features are fused via a strategy based on similarity matrices.
\begin{figure}[t]
	\centering  
	\includegraphics[width=\linewidth]{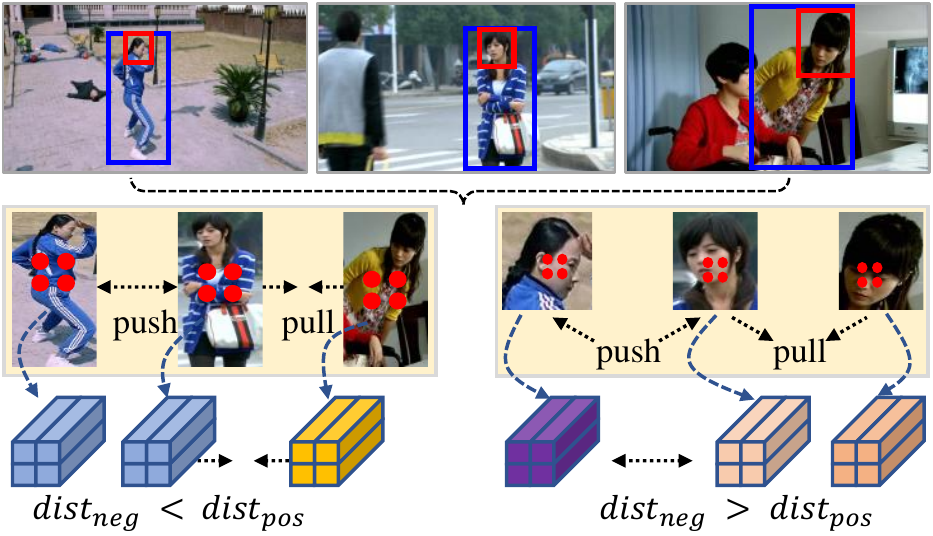}  
	\caption{\textbf{Metric learning in cloth-changing scenes.} 
		It is difficult to pull body features closer among positive cloth-changing samples, while it is much easier for head features.}	
	\label{fig:UHB}
\end{figure} 
\subsection{Unifying Head and Body Network}

As shown in Fig.~\ref{fig:framework}, the basic structure of the UHB module is based on AlignPS \cite{yan2021anchor}, which follows the one-stage anchor-free object detector FCOS \cite{tian2019fcos}. Given a whole-scene image, the UHB module aggregates features from multi-level feature maps. The largest feature map P3 is used as the output. The final objective of UHB is to learn a discriminative representation for each person box. Previous methods adopt online instance matching (OIM) loss \cite{xiao2017joint} or triplet loss \cite{wang2014learning} to supervise feature learning. However, we observe that they have some problems during training. As shown in Fig.~\ref{fig:UHB}, the woman wears a blue tracksuit and the girl changes from a blue coat to a yellow one. If only body features are considered, the distance of the negative sample pair is much smaller than that of the positive sample pair. For this case, it is very tough and also unreasonable to pull positive samples with very different colors.  

In media scenes, a considerable number of heads are available. Therefore, we design another branch for head detection and re-ID. As shown in Fig.~\ref{fig:framework}, the head branch follows the structure of the body branch. The detected results feedback to the re-ID branches. The head and body re-ID branches utilize central sampling to extract features $x$ from the P3 feature map. Both branches are optimized by the OIM loss and triplet loss, which can be denoted as: 
\begin{align}  
	&\mathcal{L}_{o} = \frac{1}{B}\sum_{i=1}^{B}\sum_{j=1}^{C}\Big[-log\frac{exp(v_j^{T}x_i/\tau)}{\sum_{k=1}^{C}exp(v_k^{T}x_i/\tau)}\Big],\\
	&\mathcal{L}_{t} = \frac{1}{B}\sum_{i=1}^{B}\Big(max\{m+d(x_i,x_i^p)-d(x_i,x_i^n), 0\}\Big),
\end{align} 
where $x_i$ is the head/body features. $B$ and $C$ are the mini-batch size and identity number. $v_j$ denotes the feature center of the $j^{th}$ identity.
$\tau$ is a softness parameter. $d(\cdot)$ denotes the distance function. $x_i^p$ and $x_i^n$ are the positive and negative features.

As shown in Fig.~\ref{fig:UHB}, the appearances of positive head pairs are much more similar than that of positive body pairs. It is reasonable to pull their distances closer, which will benefit stable training. From another point of view, the head can be regarded as a semantic part. As part-based learning has proven to be effective, the head part could enhance discriminative learning, especially in cloth-changing scenes. In the experiments, we found that high-quality heads have important impacts on final performance. Therefore, the detected heads with small confidences will be removed during training. In the inference stage,  previous part-based methods \cite{sun2018beyond} usually concatenate part features, thus yielding the failure of some headless samples. Different from them, the head and body features are used to build graphs in our method. The absence of some sample heads will not have a significant impact on training and inference. 
\begin{figure}[t]
	\centering  
	\includegraphics[width=\linewidth]{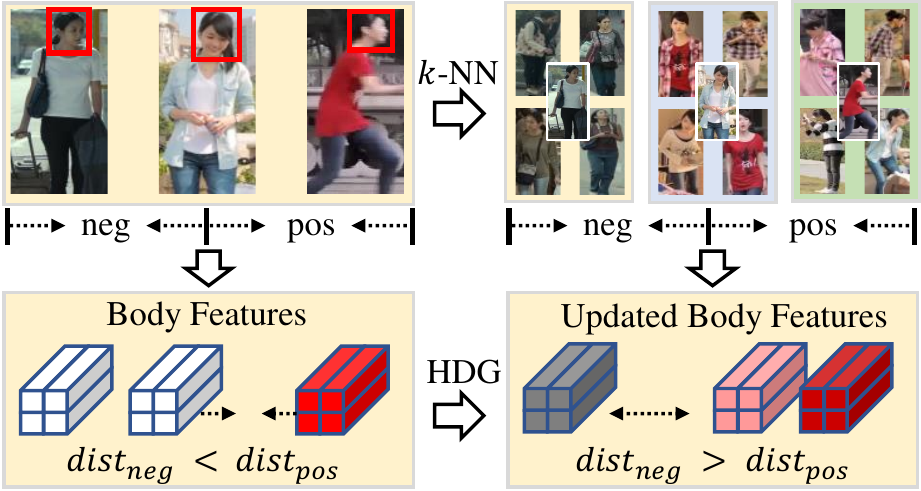}  
	\caption{\textbf{Feature updating via the HDG module.} 
		``neg'' and ``pos'' denote the negative and positive pairs. The colors denote the similarities of features. The closer the color, the more similar the features.}	
	\label{fig:HDG}
\end{figure} 

\subsection{Head-driven Graph Network}
Although the head cues are discriminative, they cannot be used for re-identification alone. This is because many heads cannot be detected or are too small to provide effective cues. To recall more positive cloth-changing samples, we further propose the head-driven graph (HDG) network. As shown in Fig.~\ref{fig:framework}, we first utilize \emph{k}-nearest neighbors (\emph{k}-NN) to build the head graph and body graph. The detected bounding boxes represent the nodes in the graph. The edges connect each node to its \emph{k} neighbors. In cloth-changing scenes, the head features are more robust than body features. Therefore, the head graph could establish more reliable links among cloth-changing samples. This is the core part of the HDG module to achieve better performance.

In the two graphs, most body and head nodes form one-to-one correlations, but still a few of them are missing. In the head graph, only the edges that both heads and bodies exist are used to update the edges of the body graph. The updated graph and body features are then input to a graph network to learn better embeddings. The graph network consists of several graph convolutional (GC) Layers. Here, we utilize the mixhop GC Layer \cite{abu2019mixhop} to explore higher-order neighbor relationships. Assume $\hat{A}$ denotes the adjacency matrix, $\hat{A}=D^{-\frac{1}{2}}(A+I_n)D^{-\frac{1}{2}}$ and $D$ is the degree matrix. All the nodes in the graph will be classified finally. The mixhop GC layer and classifier are defined as: 
\begin{align}  
	&H_{i+1} = \substack{\big|\big|\\{j\in P}}\sigma\Big(\hat{A}^j\cdot H_{i}\cdot W_{i}^j\Big),\\
	&p = f\Big(H_{i+1}\Big), 
\end{align} 
where $\hat{A}^j$ is the adjacency matrix $\hat{A}$ multiplied by itself $j$ times. $H_{i}\in\mathbb{R}^{N\times L}$ and $H_{i+1}\in\mathbb{R}^{N\times L}$ are the input and output activations of the $i^{th}$ GC layer. $N$ and $L$ denote the node quantity and feature dimension. $W_{i}^j$ is the $j^{th}$ trainable weight matrix at the $i^{th}$ GC layer. $\sigma$ is the activation function. $||$ denotes column-wise concatenation and $P$ is the set of integer adjacency powers. $f$ denotes the classifier and $p$ is the predicted probabilities. 

The cross-entropy loss is utilized to optimize the graph network. Assume $p_i$ denotes the output of the $i^{th}$ node and $y_i$ is its label. The loss function is denoted as:
\begin{equation}  
	\mathcal{L}_{g} = \frac{1}{N}\sum_{i=1}^{N}\Big(-y_i\cdot\log p_i)\Big).
\end{equation}   

To further illustrate the principle of the HDG module, we give an example in Fig.~\ref{fig:HDG}. The three images in the upper left corner form a negative sample pair and a positive sample pair. The negative pair persons both wear white jackets and dark blue jeans, but the positive person wears a red T-shirt. If body features are used directly, the distance of the negative pair is much smaller than that of the positive pair. The right corner in Fig.~\ref{fig:HDG} shows four neighbor samples for each person based on the head graph. Many of the neighbor samples have different clothes. By aggregating the neighbor features, the distance of the negative pair becomes larger than that of the positive pair. Therefore, the updated features could help retrieve more cloth-changing samples. In the inference stage, we adopt a strategy based on similarity matrices to fuse the head and updated features.
\begin{align}  
	\mathbf{S} = \lambda\cdot \mathbf{S}_{head} + (1-\lambda)\cdot \mathbf{S}_{update}, 
\end{align} 
where $\lambda$ is a weighting parameter. $\mathbf{S}_{head}$ and $\mathbf{S}_{update}$ denote the similarity matrices of the head and updated body features, respectively. The similarity matrix is the cosine similarity between the query and gallery samples. 



\begin{table}[t!]
  \centering
  \renewcommand\arraystretch{1} 
  \caption{\textbf{Comparison among PSM and previous person search datasets.}
  ``H/F'' is short for Hand/FCOS. ``FR" is short for Faster-RCNN. ``b\_boxes'' and ``h\_boxes'' denote the body and head bounding boxes, respectively.  } 
  \begin{tabular}{p{1.3cm}|p{1.2cm}<{\raggedleft}|p{1.2cm}<{\raggedleft}|p{1.2cm}<{\raggedleft}|p{1.2cm}<{\raggedleft}}

    \thickhline
    Dataset & PSM & LSPS & PRW & CUHK \\
    \hline
    \hline
    frames & 80,983 & 51,386 & 11,816 & 18,184 \\
    identities & 9,415 & 4,067 & 8,432 & 932 \\
    b\_boxes & 107,865 & 60,433 & 34,304 & 23,430 \\
    h\_boxes & 107,865 & - & - & - \\
    detectors & H/F & FR & Hand & Hand \\
    \thickhline
  \end{tabular}\label{tab:dataset}
\end{table}




\section{PSM Dataset}

\subsection{Previous Datasets}

Previous person search datasets include CUHK-SYSU \cite{xiao2017joint}, PRW \cite{zheng2017person}, and LSPS \cite{zhong2020robust}. 
CUHK-SYSU collects images from street snaps and movies, while PRW and LSPS gather images from cameras deployed on campus. Details can be found in Table \ref{tab:dataset}. As LSPS has not been released, it will not be evaluated in this work.




\subsection{Description to PSM}
Different from previous person search datasets that focus on surveillance scenes, this paper studies the media scenarios. We contribute a new large-scale dataset named Person Search in Media (PSM), which has several characteristics:

\textbf{\emph{Diversity:}} PSM covers 1,236 movies coming from more than eight countries from Asia to Europe. The person ages range from children to older people. The time spans from daytime to night. Therefore, it is very diverse in backgrounds, viewpoints, illuminations, and occlusions.

\textbf{\emph{Changing Clothes:}} In general, people in movies change clothes more frequently than in surveillance. Most people change their clothes and the maximum number of clothes attains to 24. Some even change their hairstyles, which makes PSM tougher to the person search task. The cloth-changing issue in media scenes is important and needs more studies.  

\textbf{\emph{Larger Scale:}} We give a comparison between PSM and previous datasets in Table~\ref{tab:dataset}. PSM was collected from 1,236 movies. It contains 80,983 frames and 9,415 identities, much larger than previous datasets. PSM has 107,865 body bounding boxes completely annotated manually and 107,865 head bounding boxes detected by FCOS \cite{tian2019fcos}. The head boxes are filtered and adjusted manually. To the best of our knowledge, PSM is the first person search dataset that provides head and body annotations.

\subsection{Evaluation Protocol}

PSM is split into a training set with 41,071 frames and a test set with 39,912 frames. The training and test set contain 4,755 and 4,660 identities, respectively.
Following standard person search settings \cite{xiao2017joint}, the mean Average Precision (mAP) and Rank-1 accuracy are used as evaluation metrics. The recall and average precision(AP) are employed to measure the detection performance.


\section{Experiments}
In this section, we first demonstrate the effectiveness of UDGNet on PSM dataset. Then we conduct extensive experiments to verify its generalization on surveillance datasets, other media datasets, and cloth-changing re-ID datasets. Last, we give some visualizations and discussions.
 
\subsection{Implementation Details}
Our UDGNet in Fig.~\ref{fig:framework} is implemented based on the first anchor-free person search method AlignPS \cite{yan2021anchor}. UDGNet consists of the UHB and HDG networks. To train the UHB module, the batch size of each GPU is set as 4 and the SGD optimizer is adopted. The model is trained for 24 epochs. The initial learning rate is set to 0.001 and reduced by a factor of 10 at epoch 16 and epoch 22. To train the HDG module, the learning rate is set as 0.5. The depth of the GCN module in HDG is set as 3. In our experiments, we utilize 8 NVIDIA Tesla V100 GPU to train the UDGNet, which takes around 60 hours in total.

\subsection{Comparison to the State-of-the-Arts}

\begin{table}[t!]
  \centering
  \renewcommand\arraystretch{1}
  \caption{\textbf{Comparison with recent methods on PSM dataset.} 
  ``*+UDGNet" means ``*'' is regarded as the baseline method. The proposed UDGNet can be implemented with both anchor-based and anchor-free frameworks.} 
  \begin{tabular}{p{2.6cm}|p{1.2cm}<{\centering}|p{1.4cm}<{\centering}|p{1.4cm}<{\centering}}
    \thickhline  
    \multicolumn{2}{c|}{Methods} & mAP & Rank-1 \\
    \hline
    \hline
    HOIM \cite{chen2020hierarchical} &  & 24.6 & 52.0 \\
    NAE \cite{chen2020norm} & Anchor & 33.7 & 61.9 \\
    SeqNet \cite{li2021sequential} & Based & 34.7 & 62.5 \\
    SeqNet+UDGNet & & \textbf{42.1} & \textbf{64.3} \\
    \hline
    AlignPS \cite{yan2021anchor}& Anchor & 26.2 & 51.6 \\
    AlignPS+UDGNet & Free & \textbf{38.3} & \textbf{56.5} \\
    \thickhline 
  \end{tabular}\label{tab:compinpsm}
\end{table}

 Current methods in person search can be classified into anchor-based and anchor-free ones. Anchor-free method has been newly applied to this field. As shown in Table~\ref{tab:compinpsm}, we implement our UDGNet with AlignPS \cite{yan2021anchor} and SeqNet \cite{li2021sequential}. At the same time, we evaluate other person search works on PSM based on their open-source codes.

For the anchor-free methods, the UDGNet achieves 38.3\% and 56.5\% in mAP and Rank-1 accuracy, respectively. It brings about 12.1\% gains in mAP and about 4.9\% in Rank-1 accuracy to the AlignPS \cite{yan2021anchor} method. For the anchor-based methods, the UDGNet achieves 42.1\% in mAP and 64.3\% in Rank-1 accuracy, which surpass SeqNet \cite{li2021sequential} and all other anchor-based methods. These experiments demonstrate the effectiveness of UDGNet in person search in media. 

At the same time, we observe that anchor-based methods could achieve better performance. This benefits from the two-stage detector Faster-RCNN \cite{ren2015faster}. As the anchor-free detector has been newly applied to the person search field, its performance is poorer than anchor-based methods. However, anchor-free detector has advantages in efficiency and occlusion cases \cite{tian2019fcos}. It may bring some new insights for person search and this needs more studies in the future.


\subsection{Ablation Study}

\begin{figure*}[t]
	\centering  
	\includegraphics[width=\linewidth]{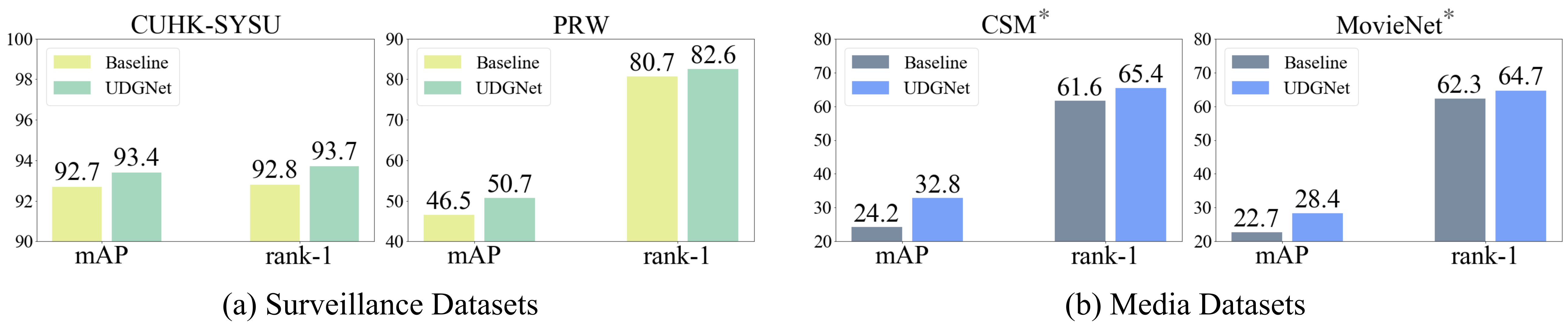}  
	\caption{\textbf{Performance comparison in surveillance and media datasets.}
		The baseline method is AlignPS \cite{yan2021anchor}. Although designed for cloth-changing scenes, UDGNet still achieves a little performance gain in CUHK-SYSU \cite{xiao2017joint} and PRW \cite{zheng2017person} datasets. It has remarkable performance gains in CSM$^*$ and MovieNet$^*$, which both provide cloth-changing cases.}
	\label{fig:sur_media}
\end{figure*}

\begin{table}
  \centering
  \renewcommand\arraystretch{1}
  \caption{\textbf{Ablation studies of different modules on PSM.} The baseline model is AlignPS \cite{yan2021anchor}.} 
  \begin{tabular}[h!]{p{1.2cm}m{1.2cm}<{\centering}m{1.2cm}<{\centering}|m{1.3cm}<{\centering}m{1.35cm}<{\centering}}
    \thickhline
    Baseline & UHB & HDG & mAP & Rank-1\\
    \hline
    \hline
    \checkmark &  & &26.2 & 51.6 \\
    \checkmark & \checkmark &  & 32.9 & 55.5 \\ 
    \checkmark & \checkmark & \checkmark  & \textbf{38.3} & \textbf{56.5} \\
    \thickhline

  \end{tabular}\label{tab:abla}
\end{table}



\begin{table}[t!]
  \centering
  \renewcommand\arraystretch{1}
  \caption{\textbf{Comparison between the body-driven graph and head-driven graph.}} 
  \begin{tabular}{p{3.7cm}|p{1.7cm}<{\centering}|p{1.7cm}<{\centering}}
    \thickhline
    Methods & mAP & Rank-1 \\
    \hline
    \hline
    Body-driven Graph & 36.4 & 55.5\\
    Head-driven Graph & \textbf{38.3} & \textbf{56.5} \\
    \thickhline
  \end{tabular}\label{tab:graph}
\end{table}
 
\begin{table}[t!]
  \centering
  \renewcommand\arraystretch{1}
  \caption{\textbf{Comparison of detectors between the baseline and UDGNet.} The head detector has a little but trivial impact on the body detection.} 
  \begin{tabular}{p{1.8cm}|p{1.6cm}<{\centering}|p{1.6cm}<{\centering}|p{1.6cm}<{\centering}}
    \thickhline
    Methods & Detection & AP & Recall \\
    \hline
    \hline
    AlignPS & body & 67.4 & 94.3 \\
    UDGNet & body & 67.3 & 93.4\\
    UDGNet & head & 63.1 & 93.7\\
    \thickhline
  \end{tabular}\label{tab:det}
\end{table}

\textbf{Effectiveness of UHB} As shown in Table~\ref{tab:abla}, the UHB module achieves 32.9\% in mAP and 55.5\% in Rank-1 accuracy, respectively. It has improved the performance of baseline method significantly. The performance gain benefits from the joint optimization of the head branch. Note that only the body feature from the body branch is used in the test stage, which consistences with the baseline method. Since two branches(body and head) share the same backbone, the optimization of the head branch also benefits the body branch. That is to say, the head cue plays an important role in cloth-changing scenarios. It guides the feature extractor focus more on the discriminative parts. 

\textbf{Effectiveness of HDG } Table~\ref{tab:abla} shows us that the HDG module further improves the mAP from 32.9\% to 38.3\%. This indicates that HDG recalls more hard positive samples, which are usually cloth-changing samples. The HDG module first establishes reliable links via the head graph, then aggregates the neighbour features to pull the distances of positive cloth-changing samples. This operation is the key for HDG to achieve performance improvement. The experiments not only verify the effectiveness of HDG, but also show us the importance of linking cloth-changing samples.



\textbf{Why head graph is needed?} In cloth-changing scenes, the head graph is considered to be able to establish more reliable links than the body graph. 
To verify the effectiveness, we conduct experiments in Table~\ref{tab:graph}. The head-driven graph shows better results on both mAP and Rank-1 accuracy compared with the body-driven graph. That indicates the effectiveness of the head-driven graph. Considering that we use \emph{k}-NN to build the graph, the accuracy of neighbor nodes largely affects the quality of the graph and further influences the final performance. In cloth-changing scenes, the clothing cues are not reliable enough. Body features tend to build links among the same clothes, while head features are more likely to link cloth-changing samples. Therefore, the head-driven graph can achieve better performance.

\subsection{Evaluation of Detection}
We evaluate the detection performance between AlignPS and UDGNet in Table~\ref{tab:det}. The AlignPS means only the body is detected and UDGNet denotes that the body and head are detected simultaneously. 
It shows us that the head detector has a little but trivial impact on the body detector. 
Compared with Table~\ref{tab:abla}, we have an interesting observation: the re-ID performance is improved with a large margin, although the detection performance has a slight drop. This is due to the conflict between person detection and re-ID. Person detection deals with common appearances, while re-ID focuses on a person's uniqueness. Based on the unified detector, we do not need to pre-train a head detector, which has significant advantages in efficiency.




\subsection{Evaluation in Surveillance}
In this part, we evaluate UDGNet in current person search datasets, \emph{i.e.,} CUHK-SYSU \cite{xiao2017joint} and PRW \cite{zheng2017person}. As they do not provide head bounding boxes, we utilize FCOS \cite{tian2019fcos} to pre-detect the heads and only keep the head boxes with high centerness scores. Experimental results are shown in Fig.~\ref{fig:sur_media}(a). In CUHK-SYSU, UDGNet achieves 93.4\% in mAP and 93.7\% in Rank-1, which are slightly higher than AlignPS. In PRW, UDGNet achieves 50.7\% in mAP and 82.6\% in Rank-1 accuracy. It brings about 4.2\% mAP gains and 1.9\% Rank-1 accuracy improvements. These results indicate that our method can still achieve satisfying performance when applied to surveillance datasets. Noting that the performance improvement in PRW is higher than CUHK-SYSU. We carefully analyzed the experiments and found that each query in CUHK-SYSU has less than two positive samples in the gallery. However, each query in PRW has 37 positive samples on average. The small number of positive samples in the gallery cannot reflect the advantages of the proposed method.


\subsection{Evaluation on Media Datasets}
CSM \cite{huang2018unifying} and MovieNet \cite{huang2020movienet} are not specifically designed for person search task, but for video understanding. Since they contain identities and body bounding boxes, they can also be used for person search. It should be noted that the query images they provided are portraits, not complete human bodies, leading to the unavailability of body cues. Besides, the provided samples are tracklets with highly redundant frames. Therefore, we uniformly sample frames and get two subsets named CSM$^*$ and MovieNet$^*$. CSM$^*$ has 845 identities, 12,638 frames, and 500 query boxes. MovieNet$^*$ has 2,592 identities, 63,809 frames, and 1,000 query boxes. We ensure that each query identity appears at least four times in the gallery. 
The experimental results on CSM$^*$ and MovieNet$^*$ are present in Fig.~\ref{fig:sur_media}(b). It is clearly shown that UDGNet has achieved a gain of 8.6\% and 5.7\% in mAP, respectively. This is a significant performance improvement compared with the baseline method. Such improvements verify the generalization and effectiveness of the proposed UDGNet in media scenarios.

\subsection{Evaluation on Cloth-Changing re-ID Datasets}

  \begin{table}
  \centering
  \renewcommand\arraystretch{1}
  \caption{\textbf{Experimental Results on cloth-changing re-ID Datasets.} 
  The baseline method is BoT \cite{luo2019bag}. As the re-ID datasets are cropped images, we remove the UHB module and only evaluate the HDG module.}
  \begin{tabular}[h!]{p{1.6cm}|p{1.05cm}<{\centering}|p{1.1cm}<{\centering}|p{1.1cm}<{\centering}|p{1.3cm}<{\centering}}
    \thickhline
    \multicolumn{5}{c}{Celeb-reID} \\
    \hline
    Methods & mAP & Rank-1 & Rank-5 & Rank-10 \\
    \hline
    \hline
    BoT &  11.0 & 54.3 & 68.0 & 75.0\\
    BoT+HDG & \textbf{14.5} & \textbf{57.0} & \textbf{74.2} & \textbf{79.5} \\ 
    \hline
    \hline
    \multicolumn{5}{c}{PRCC} \\
    \hline
    Methods & mAP & Rank-1 & Rank-5 & Rank-10 \\
    \hline
    \hline
    BoT & 30.3 & 32.2 & 41.4 & 45.6 \\
    BoT+HDG & \textbf{37.7} & \textbf{50.4} & \textbf{63.3} & \textbf{66.1} \\
    \thickhline
  \end{tabular}\label{tab:reid}
\end{table}

Although this work focuses on person search in media, we are still interested in the performance of the proposed method in cloth-changing re-ID datasets, \emph{i.e.,} Celeb-reID \cite{huang2019celebrities} and PRCC \cite{yang2019person}. As re-ID datasets are cropped images, we remove the UHB module and only evaluate the HDG module in this section. First, the FCOS \cite{tian2019fcos} detector is used to detect the heads of Celeb-reID and PRCC datasets. Then, we crop the heads based on the detected bounding boxes and regard them as training samples. Next, we use BoT \cite{luo2019bag} to train the body samples and head samples independently. Finally, the extracted head and body features are used as the input of the HDG module.
 
Experimental results are shown in Table~\ref{tab:reid}. In Celeb-reID, the HDG module has achieved 14.5\% in mAP and 57.0\% in Rank-1 accuracy, which is better than the baseline method. In PRCC, the HDG module achieves a performance improvement of 7.4\% in mAP and 18.2\% in Rank-1 accuracy, respectively. These experiments fully demonstrate the effectiveness of the proposed HDG module in cloth-changing scenes.

\subsection{Visualization}

\begin{figure}[t]
	\centering
	\includegraphics[width=\linewidth]{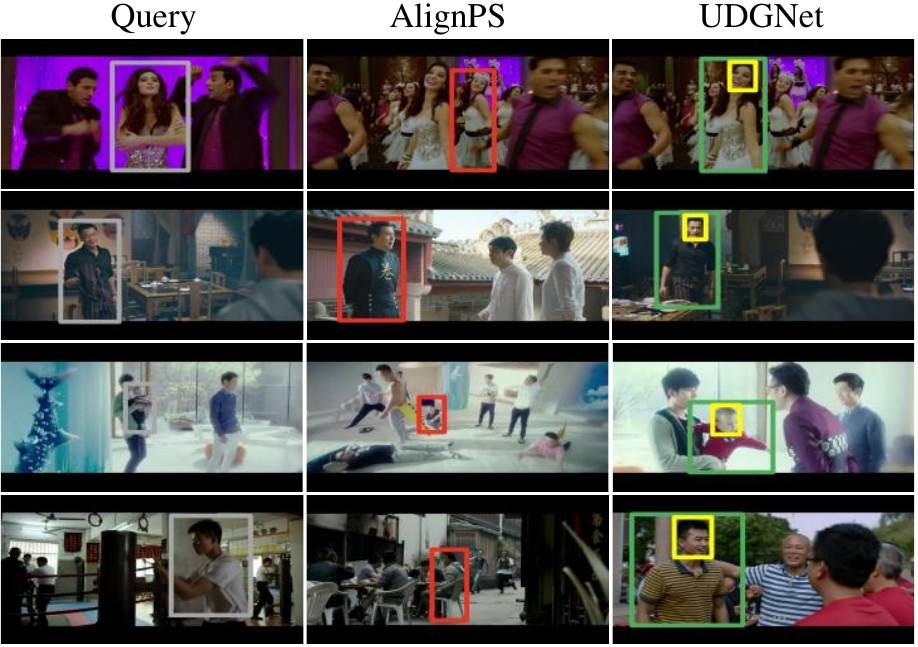} 
	\caption{\textbf{The top-1 detection and re-identification on PSM dataset.}
	The grey boxes indicate the queries. The green/red boxes denote the true/false results. The yellow boxes are the detected heads matched with detected bodies.
	}
	\label{fig:searchvis}
\end{figure}

\textbf{Top-1 detection and re-identification.} As shown in Fig.~\ref{fig:searchvis}, our model could accurately detect the human body and head simultaneously. This is the foundation for person search task. The first and second lines give cases in which clothes are not changed. The third and fourth lines show cloth-changing cases. It shows us that AlignPS tends to locate persons who have similar appearances. Once other persons wear similar clothes or the target changes clothes, the model may make mistakes. Benefitting from the head cues, UDGNet could locate the correct targets in these scenes, even if they have different clothes with the query image.

\begin{figure}[t]
	\centering
	\includegraphics[width=\linewidth]{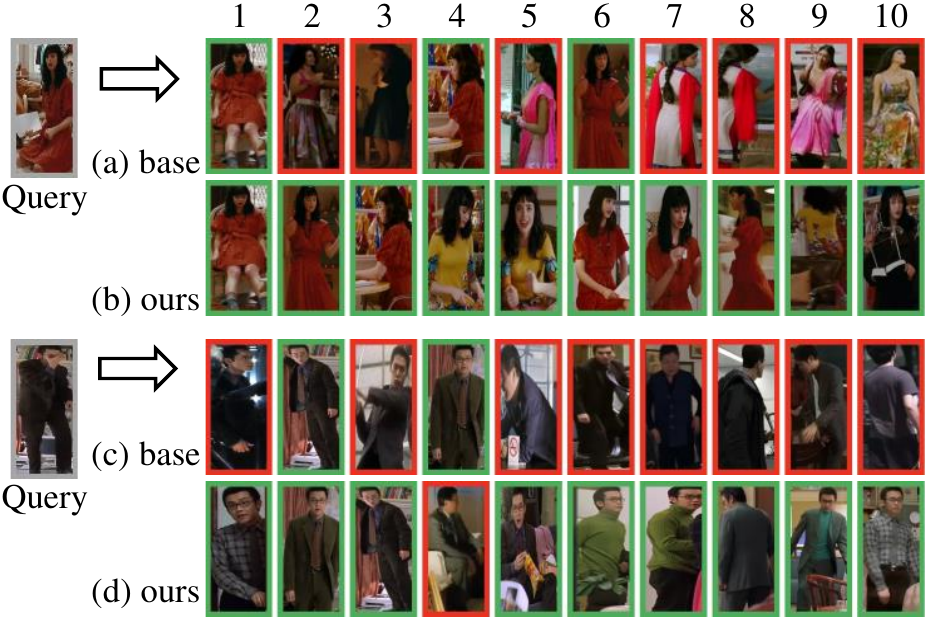} 
	\caption{\textbf{The top-10 retrieved results on PSM dataset.}
	The red color boxes denote failure cases, and the green color boxes denote right cases.
	}
	\label{fig:reid_visual}
\end{figure}

\textbf{Top-10 retrieved results.} To further verify whether our model could recall more cloth-changing samples, the top-10 retrieved results are shown in Fig.~\ref{fig:reid_visual}. The query girl is wearing a red dress. The baseline method can retrieve some positive samples, but all of them are the same clothes. Our method has retrieved more positive samples, which contain four cloth-changing samples. Specifically, the top-9 sample cannot see the face, but our model still retrieves it. This benefits from the head cue, \emph{e.g.,} the hairstyle. The query man is wearing a dark suit. For the baseline method, almost all the retrieved samples are dark clothes. However, our model could retrieve much more cloth-changing positive samples.


\subsection{Results \& Analysis}
This work studies the cloth-changing issue in person search in media. By extensive experiments and analysis, we have demonstrated the effectiveness of UDGNet in cloth-changing scenes and found two important answers: 1) The head detection and re-identification could effectively enhance discriminative learning. 2) The head graph plays an important role in recalling positive cloth-changing samples. In this work, we utilize the head cue to establish reliable links among cloth-changing samples. Maybe more related cues can be studied in the future. It is a promising direction to further improve the performance and enhance the robustness of the person search framework.
  

\section{Conclusion}
This paper proposes a unified UDGNet framework for person search in media. UDGNet simultaneously conducts head/body detection and re-identification based on the same network. Besides, it builds a head-driven graph network to address the cloth-changing issue. This paper also contributes a PSM dataset, which is by far the largest person search dataset in media. Extensive experiments demonstrate that UDGNet can achieve satisfying performance in cloth-changing scenes, including PSM, CSM, MovieNet, PRCC, and Celeb-reID. Moreover, UDGNet can be implemented with both anchor-free and anchor-based frameworks and further bring considerable performance improvement.
   

{\small
\bibliographystyle{ieee_fullname}
\bibliography{egbib_short}
}

\end{document}